# ANY TIME PROBABILISTIC REASONING FOR SENSOR VALIDATION

**P.H. Ibargüengoytia**
Instituto de Investigaciones
Eléctricas, A.P. 1-475
Cuernavaca, Mor., 62001, México
pibar@iie.org.mx

**L.E. Sucar**
Instituto Tecnológico y de
Estudios Superiores de Monterrey
Campus Morelos, A.P. 99-C
Cuernavaca, Mor., 62050, México
esucar@campus.mor.itesm.mx

**S. Vadera**
University of Salford
Dept. of Mathematics
and Computer Science
Salford, M5 4WT, U.K.
S.Vadera@mcs.salford.ac.uk

## Abstract

For many real time applications, it is important to validate the information received from the sensors before entering higher levels of reasoning. This paper presents an any time probabilistic algorithm for validating the information provided by sensors. The system consists of two Bayesian network models. The first one is a model of the dependencies between sensors and it is used to validate each sensor. It provides a list of potentially faulty sensors. To isolate the real faults, a second Bayesian network is used, which relates the potential faults with the real faults. This second model is also used to make the validation algorithm any time, by validating first the sensors that provide more information. To select the next sensor to validate, and measure the quality of the results at each stage, an entropy function is used. This function captures in a single quantity both the certainty and specificity measures of any time algorithms. Together, both models constitute a mechanism for validating sensors in an any time fashion, providing at each step the probability of correct/faulty for each sensor, and the total quality of the results. The algorithm has been tested in the validation of temperature sensors of a power plant.

## 1 Introduction

Artificial intelligence (AI) techniques are playing an increasingly important role in real applications. In industry, different techniques have been proposed, for example, in diagnosis, automatic control, and monitoring. Generally, these applications require an overall model which usually, its inputs are mainly sensors. Also, many of these real applications need to maintain a real time behaviour, i.e., the correctness of the system depends not only on the logical result of the computation but also on the time at which the results are produced [Stankovic 1988]. Usually, real applica-tions possess a time limit by which some actions must be performed.

This paper presents a model for the validation of the sensors used in real time applications. The proposed validation is carried out in a separate module that works together with other functions in a system. In other words, it is assumed that a layered scheme is used in which the lowest level concentrates on validating the signals transmitted by the sensors as presented in Fig. 1 [Yung & Clarke 1989]. The main benefit of using a layered approach is that it enables the construction of models in a modular fashion. That is, it is easier to construct a model for sensor validation and then a model for the process than it is to construct an overall model in one step. This separation of the sensor validation layer can also result in simpler higher layer models and leave the higher layers to utilize other techniques.

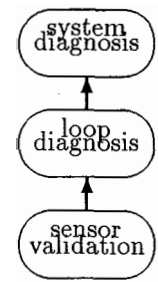

Figure 1: Layered diagnosis architecture.

Faults in the sensors readings are detected in a decentralised and hierarchical approach, so that they can be easily isolated and repaired. Additionally, suppose that the higher layers of the system represent other important and critical functions, e.g., the fault diagnosis of a nuclear plant. The intermediate layer (loop diagnosis) may be using *model based reasoning* to diagnose a control loop in the plant, whereas the system diagnosis layer may be utilizing a different approach. The validation module presented in this paper, utilizes a probabilistic model which considers only the relationships between the variables to be validated. This probabilistic model is independent of the higher layers models, so it is easier to construct and mantain when necessary.

This paper presents the continuation of the project described in a previous paper [Ibargüengoytia et al. 1996]. In that paper, the authors described a probabilistic approach to sensor validation that took advantage of a Markov blanket property to distinguish real faults from apparent faults. A Bayesian network was used as a basis for predicting a probability distribution for a sensor value based on other sensors. The predicted distribution and the actual sensor reading was used in order to determine if there was a potential fault.

However, the sensor validation process described in that paper works in batch mode, i.e., no intermediate results are available, and no attempt is made to estimate the quality of the results. For a real time application, these characteristics are inadequate. This paper presents the extension of the sensor validation algorithm, so it can be applied in real time systems. This consists in the use of *any time algorithms*.

Thus, the extension of the sensor validation algorithm consists in the following features. First, the use of a probabilistic causal network that relates the real and apparent faults. Second, in order to perform in any time basis, the validation algorithm selects the sensor which provides more information when validated. Finally, a quality function is calculated in order to characterize the behaviour of the algorithm. The selection of the most *informative* sensor is made using the entropy function.

To summarize, this paper presents an any time probabilistic algorithm for validating the information provided by sensors. The system consists of two Bayesian network models. The first one is a model of the dependencies between sensors and it is used to validate each sensor. It provides a list of potentially faulty sensors. To isolate the real faults, a second Bayesian network is used, which relates the potential faults with the real faults. This second model is also used to make the validation algorithm any time, by validating first the sensors that provide more information. To select the next sensor to validate, and measure the quality of the results at each stage, an entropy function is used. Together, both models constitute a mechanism for validating sensors in an any time fashion, providing at each step the probability of correct/faulty for each sensor, and the total quality of the results.

The next section briefly describes the basis of any time algorithms.

## 2   Any Time Algorithms

Any time algorithms represent one direction of work that aims to achieve the use of artificial intelligence techniques in real time systems. This term was initially used by Dean in his research about time dependent planning [Dean & Boddy 1988]. At the same time, Horvitz (1987) proposed the name of *flexible computation* for this mechanism. Any time algorithms are those that can be interrupted at any point during computation, and return an answer whose *value* increases as it is allocated additional time [Boddy & Dean 1994]. However, how can this value be measured in a specific application? The literature contains descriptions of different dimensions that have been proposed as metrics [Zilberstein & Russell 1996]: certainty, accuracy and specificity.

*Performance profiles* represent the expected value of these metrics for a given procedure as a function of time. In other words, performance profiles characterize the quality of an algorithm's output as a function of computation time. Figure 2 illustrates three cases of performance profiles [Zilberstein & Russell 1996], [Dean & Boddy 1988]:

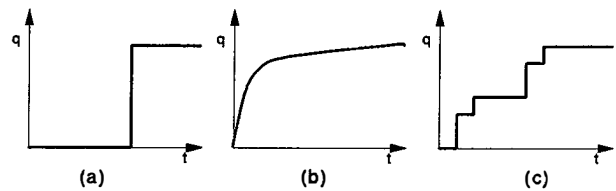

Figure 2: Examples of performance profiles. (a) a standard or one shot algorithm. (b) an ideal, exponential precision algorithm, and (c) a more realistic profile for an any time algorithm in practice.

Clearly, all these types of performance profiles are special cases of a superclass that can be defined as *monotonic improvement*, i.e., the quality of its intermediate results does not decrease as more time is spent to produce the result. The next section explains the basis of the validation algorithm, so that section 4 develops the any time algorithm for the sensor validation problem.

## 3   Sensor Validation

The probabilistic sensor validation model utilizes Bayesian networks. The nodes represent the measures of the sensors. The structure of the network makes explicit the dependence relations between the variables.

The probabilistic sensor validation includes the diagnosis of all single sensors in the network. The idea is to instantiate all the nodes with the sensor readings, except the one being validated. A probabilistic propagation provides a distribution of the posterior probability of the estimation of a signal value based on the readings of the most related signals. The estimated value is then compared with the current value in order to decide if the measurement is correct. The most closely related variables for each sensor consist of a *Markov blanket* of the sensor variable. A Markov blanket is defined as the set of variables that makes a variable independent from the others. In a Bayesian network, the following three sets of neighbours is sufficient for forming a Markov blanket of a node: the set of direct predecessors, direct successors, and the direct predecessors of the successors (i.e. parents, children, and spouses) [Pearl 1988]. The set of variables that constitutes the Markov blanket of a variable can be seen as a protection of this variable against changes of variables outside the blanket. This means that, in order to analyze a variable, it is only necessary to know the value of all variables in its blanket [Ibargüengoytia et al. 1996]. Additionally, the *extended Markov blanket* (EMB) of a sensor is formed by its Markov blanket plus the variable itself.

However, since validating a sensor based on a faulty one results in an erroneous validation, the probabilistic validation only provides a list of apparent faults. Thus, the probabilistic validation provides a list of potential correct and potential faulty sensors. The fault isolation is carried out when the list of potential faulty sensors is compared with the list of EMB of each sensor. When a match exists, then the faulty sensor has been distinguished. Otherwise, different conditions exist for the isolation of multiple failures [Ibargüengoytia et al. 1996]. The next section describes the extensions of the sensor validation model in order to discriminate faulty and correct sensors in an any time basis.

## 4    Any Time Sensor Validation

Any time sensor validation algorithm implies that the knowledge about the state of the sensors (faulty or correct) becomes more certain and complete as time progresses. Certainty about the state of a sensor refers to the degree of belief in the correctness of a sensor, and completeness is characterized by the number of sensors from which the state is known. Thus, it is required to be able to monitor the state of the sensors during all the validation process. This is done through a vector whose elements $P_f(s_i)$ represent the probabilities of failure for the sensors $s_i$. Given that the any time validation process needs to be cyclic, the top level of the algorithm can take the form shown in Fig. 3.

---

1. Initialize $P_f(s_i)$ for all sensors $s_i$.
2. While there are unvalidated sensors do:
   (a) choose the next sensor to validate
   (b) validate it
   (c) update the probability of failure vector $P_f$
   (d) measure the quality of the partial response

---

Figure 3: Top level of the any time sensor validation algorithm.

The probabilistic validation of a single sensor (step b)

will be explained next in order to clarify the rest of the algorithm.

### 4.1    Validation

The validation step was briefly introduced in section 3 and more extensively in [Ibargüengoytia et al. 1996].

The sensors are processed one by one by the validation function utilizing the following algorithm:

1. Read the actual value of the variable provided by the sensor.

2. Read the value of all variables that appear in the Markov blanket of the selected variable.

3. Propagate the probabilities and obtain the posterior probability distribution for the selected variable.

4. If the probability (obtained in 3) of the value acquired in step 1 is lower than a specified value, return *failure*; else return *success*

For example, consider the simplified model of a gas turbine shown in Fig. 4. The validation of $m$ is carried out by calculating the probability distribution of $m$ given the measurements of $t$ and $p$. If the real value of sensor $m$ has a probability greater than certain value, then the sensor is considered correct, and faulty otherwise. However, if the fault is in sensor $p$, then the validation of $m$ will also result in apparent fault.

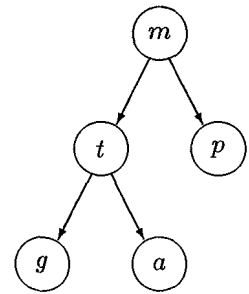

Figure 4: A reduced Bayesian network of a gas turbine.

Thus, the validation step is carried out by this algorithm that receives as input, the sensor that will be validated. As output, the algorithm returns a binary value {correct,faulty} with the apparent status of the sensor.

### 4.2    Selection of next sensor

This section develops a mathematical model for choosing the best sensor to validate given the history of the validation process and the current state of the system. Also, the model proposed here will be used for measuring the quality of the response in order to obtain the performance profile of the validation algorithm. The central idea is that the validation of a sensor provides some information and also, extra information can be inferred. Therefore, a measure of the information that a single validation produces is required. A definition of the expected amount of information that an event produces was first proposed by Shannon and used in communication theory [Shannon & Weaver 1949]. Shannon proposed the following definitions.

**Definition 4.1** *Given a finite probability distribution*

$$p_i \geq 0 \; for \; (i = 1, \ldots, n), \; and \; \sum^n p_i = 1$$

*Shannon's entropy measure is defined as*

$$H_n = H_n(p_1, \ldots, p_n) = -\sum_{i=1}^{n} p_i log_2 p_i \qquad (1)$$

Thus, the entropy measures the related number of bits required to store the information.

Since the validation of a sensor $s$ has two possible outcomes, the entropy function $H(s)$ is then defined as:

$$H(s) = \begin{cases} 0 & \text{if } p = 0 \text{ or } p = 1 \\ -plog_2(p) - (1-p)log_2(1-p) & otherwise \end{cases} \qquad (2)$$

where $p$ represents the probability of failure of the sensor. Notice that the expression $plog_2(p) = 0$ when $p = 1$ but it is undefined when $p = 0$. However, since $plog_2(p)$ tends to zero as $p$ tends to zero, the values defined in equation 2 can be safely assumed. Notice that it has its maximum when $p = \frac{1}{2}$, i.e., when the ignorance is maximum, and it is zero when either $p = 0$ or $p = 1$, i.e., when the information is maximum and ignorance is minimum. This function can be considered either as a measure of the information provided by an experiment, or as a measure of the uncertainty in the experiment's outcome. Thus, considering each single sensor validation as an experiment, this function can be used to measure the amount of information provided by that validation. Then, the average amount of information $\mathcal{E}$ for the system can be defined as follows:

$$\mathcal{E}(s_1, \ldots, s_n) = \frac{1}{n} \sum_{i=1}^{n} H(s_i)$$

$$= -\frac{1}{n} \sum_{i=1}^{n} P_f(s_i) log_2 P_f(s_i)$$

$$+ (1 - P_f(s_i)) log_2(1 - P_f(s_i))$$

$$= -\frac{2}{n} \sum_{i=1}^{n} P_f(s_i) log_2 P_f(s_i) \qquad (3)$$

where $n$ is the number of sensors in the system $S$, and $P_f(s_i)$ represents the current probability of failure value assigned to sensor $s_i$. Notice that the vector whose elements are $P_f(s_i)$ provides a measure of the certainty in the validation while the sum of $n$ individual entropies provides a specificity measure of the result.

Given this measure, the any time sensor validation algorithm needs to select a sensor $X$ that gives the best improvement in the average entropy of the system $S$. Hence the following conditional version of equation 3 can be written

$$\mathcal{E}(S \mid X) = \mathcal{E}(S \mid x = ok) + \mathcal{E}(S \mid x = flty)$$

$$= \frac{1}{n} \left( \sum H(s_i \mid x = ok) + \sum H(s_i \mid x = flty) \right) \qquad (4)$$

This function can be evaluated for each sensor and the one which gives the most information (the minimum $\mathcal{E}(S \mid X_i)$) can be selected as the next sensor $X_i$ to be validated. The computation suggested by the above formulae could be too expensive for a real time sensor validation process. To overcome this problem, a pre compilation of the sensor selection mechanism is implemented as follows. The above formulae are used to select the sensor, $s_r$ which gives the most information. This selected sensor forms the root of a binary decision tree. A fault is simulated in this sensor and the formulae are again used to select the next sensor $s_{r-}$. Then, the root $s_r$ is assumed to be correct, and the formulae are used to select the sensor $s_{r+}$ in this case. This results in the partial decision tree shown in Fig. 5. This process is repeated recursively on the nodes $s_{r-}$ and $s_{r+}$ to obtain a complete decision tree, so that each path in the tree includes all the sensors.

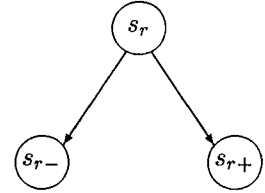

Figure 5: Partial decision tree.

As an example, consider the network shown in Fig. 4. This process results in the decision tree shown in Fig. 6. Notice that this tree can be reduced considering only the valid trajectories formed by the assumption of, for example, single faults among the set of sensors. See [Ibargüengoytia 1997] for more details.

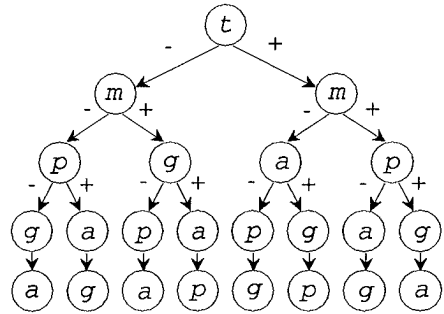

Figure 6: Binary tree indicating the order of validation given the response of the validation step.

This decision tree can be used to select the next sensor more efficiently in real time than by performing the calculations. Thus, the selection step of the algorithm of Fig. 3 consists of simply traversing the tree one level after every single sensor validation. The cycle starts at the root, and the decision tree points to the next node in the tree according to the result of validating the current sensor.

### 4.3  Isolation

The validation step provides only a list of potentially faulty sensors. Thus, a comparison is made between the set of potentially faulty sensors with the table of extended Markov blankets of all the sensors. When a match is found, a real fault is determined. However, the set of potentially faulty sensors is obtained after all the sensors have been validated. Therefore, in order to extend that algorithm for any time behaviour, a different mechanism for distinguishing real faults from apparent ones is required. This new mechanism provides, as the output of the isolation phase, a vector with the probability of a real fault in all the sensors. This vector is refined incrementally in time, so the any time behaviour can be achieved.

The any time fault isolation process is based on the relationship between real and apparent faults. There are two situations that arise: (i) the existence of a real fault *causes* an apparent fault (as shown in Fig. 7(a)), and (ii) one apparent fault is the manifestation of several possible real faults (as shown in Fig. 7(b)).

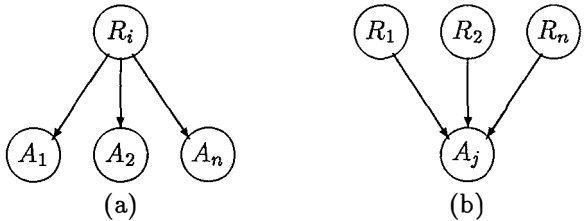

Figure 7: Causal relation between real faults ($R$) and apparent ($A$) faults represented as nodes. In (a), one real fault causes several apparent ones, while in (b), one apparent fault is caused by one or more real faults.

In both figures, the relation between root nodes and leaf nodes is the same as the extended Markov blanket (EMB) of a sensor. Considering all the sensors, a causal model relating the real and apparent faults can therefore be obtained from the fault detection Bayesian network (in fact, the EMB table is sufficient to build this network). In the first level (roots), the nodes represent the events of real failure in every sensor. Then, the second level (leaves) is formed by nodes representing apparent failures in all the sensors. Arcs are included between every root node, and the corresponding nodes of the extended Markov blanket. For example, the causal network shown in Fig. 8 can be obtained directly from the Bayesian network of the gas turbine given in Fig. 4. Thus, the consequences of observing an apparent fault can be propagated in the causal network in order to obtain the probabilities of a real fault in all the sensors.

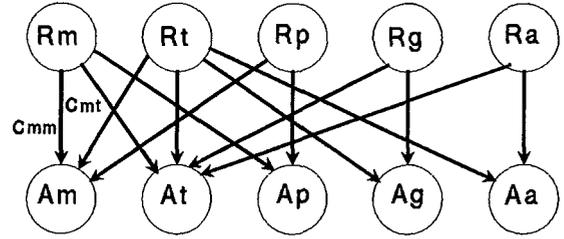

Figure 8: Probabilistic causal model for fault isolation in the example of Fig. 4. $R_i$ represents a real fault in sensor $i$ while $A_j$ represents an apparent fault in sensor $j$.

The network of Fig. 8 is multiply connected. Hence, the propagation method of trees of cliques is utilized [Lauritzen & Spiegelhalter 1988].

In general, $O(2^n)$ conditional probabilities would be required (for a node with $n$ parents). However, the *noisy or* model can be adopted here. Two assumptions need to hold in order to use this model: accountability and exception independence [Pearl 1988].

The accountability assumption holds by the way the model is constructed, i.e., a sensor is apparently faulty only if there is a fault in its MB. The exception independence assumption is concerned about a rare situation for this particular model. The relationship between the real and apparent faults is obtained from a Bayesian network in which the dependencies are assumed to be *strong*. Hence, the probability of a real fault not resulting in an apparent fault is small. Further, the mechanism by which a real fault in one sensor does not result in an apparent fault is even less likely to be dependent on another real fault. Hence, given that these assumptions are reasonable, the conditional probability matrix can be calculated by utilizing equation 5.

$$P(A_j \mid \mathbf{d}) = \begin{cases} \prod_{i \in T_d} q_{ij} & \text{if } \neg A_j \\ 1 - \prod_{i \in T_d} q_{ij} & \text{if } A_j \end{cases} \quad (5)$$

where $\mathbf{d}$ is the set of assignments of the set of apparent faults, and $T_d$ represent the set of all apparent faults actually present. Thus, the only parameter required is defined as:

$$c_{ij} = 1 - q_{ij} = P(A_j \mid R_i \ only).$$

In the case of the sensor validation problem, in an ideal case, all the parameters $c_{ij} \approx 1$. Of course, these values can be obtained by simulation from the data if the problem is expected to depart from this ideal case. That is, according to the theory developed in Ibargüengoytia et al. (1996), when a real fault $R_i$ is present, it will always cause the apparent fault $A_j$ (assuming that there is an arc from $R_i$ to $A_j$).

The network of Fig. 8 is initialized with the following information: (i) the prior probability of all the root nodes in the model is 0.5 (assuming ignorance at the beginning of a cycle), and (ii) the parameters $c_{ij} = 0.99$ for all $1 \leq i, j \leq$ number of nodes.

Having described how real and apparent faults can be related, the fault isolation model can now be summarized. It receives as an input, a validated sensor with its detected state (faulty or correct) and updates the probability of failure of all the sensors. It does this by instantiating the value of the corresponding apparent node and using a propagation algorithm to obtain the posterior probabilities of the real faulty nodes. A vector $P_f$ of these posterior probabilities represents the current state of knowledge about the sensors, and can be viewed as the output of the system at any time. For example, assuming a fault in $g$ in the network of Fig. 4, produces the sequence of values of the probability vector as shown in Table 1.

Table 1: Example of the values of the probability vector $P_f$.

| Step | $P_f(m)$ | $P_f(t)$ | $P_f(p)$ | $P_f(g)$ | $P_f(a)$ |
|------|----------|----------|----------|----------|----------|
| $t = faulty$ | 0.534 | 0.534 | 0.5 | 0.534 | 0.534 |
| $m = correct$ | 0.013 | 0.013 | 0.009 | 0.663 | 0.663 |
| $g = faulty$ | 0.009 | 0.019 | 0.009 | 0.99 | 0.502 |
| $a = correct$ | 0.009 | 0.0 | 0.009 | 0.999 | 0.009 |
| $p = correct$ | 0.0 | 0.0 | 0.0 | 0.999 | 0.009 |

### 4.4  Quality measure

A measure that is independent of the application is the average entropy of the sensors given in equation 3. That is, if the current quality measure is:

$$Q(s_1, \ldots, s_n) = -\frac{2}{n} \sum_{i=1}^{n} P_f(s_i) log_2 P_f(s_i) \qquad (6)$$

then, the reported quality function is calculated with the formula $Q = \frac{Q_{max} - Q_{current}}{Q_{max}}$ where $Q_{max}$ is the maximum value of the quality measure (i.e., $n$, the number of nodes). Notice that this measure captures both the certainty and specificity measures of any time algorithms. It captures certainty since the probabilities of the sensors are used, and specificity since all the sensors are combined to give an average. Figure 9 shows the performance profile obtained with this quality measure for the example of Fig 4.

## 5  Empirical Results

The sensor validation algorithm was evaluated by applying it to the validation of temperature sensors of the gas turbine at the *Gómez Palacio* power plant in

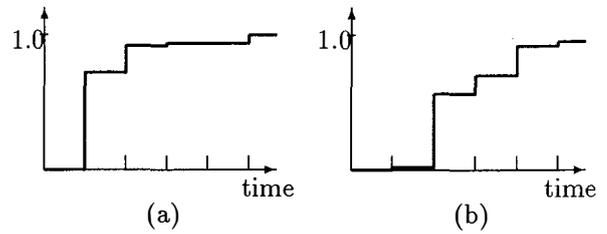

Figure 9: Performance profile describing the combination of certainty and specificity in one parameter against time. (a) without failure, (b) with a simulated failure in sensor $g$.

México. A Bayesian network representing the *dependencies* between the sensors of the plant is shown in Fig. 10. The dependency model was obtained by utilizing an automatic learning program that uses real data from the start up phase of the turbine [Sucar et al. 1997].

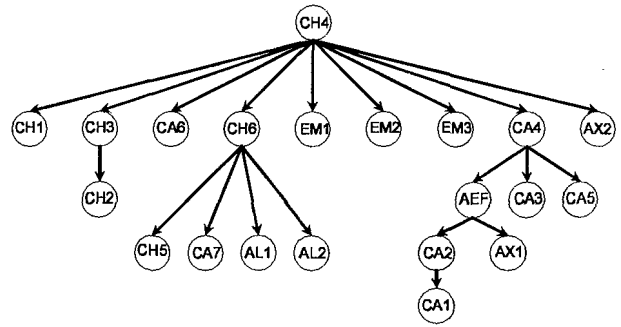

Figure 10: Probabilistic tree of the application. Nodes represent temperature signals of a gas turbine.

The data set was partitioned in two subsets: one partition for training the network, and the other partition for testing. The training/testing partition used was 70-30 % of the original data set, i.e., 610 instances for training the model (calculating the prior and conditional probabilities), and 260 instances for testing.

Theoretically, the system should always detect and isolate single faults correctly. However, in reality, some errors may occur since in practice it is unlikely that the dependency model will be perfect. Consequently, two types of errors could occur: a correct reading might be considered faulty, and a real fault might not be detected. These two possible errors are called type I and type II errors in the literature, and defined as follows [Cohen 1995]:

**type I:** rejection of the *null hypothesis* when it is true,

**type II:** acceptance of the *null hypothesis* when it is actually false.

The *null hypothesis* used refers to the hypothesis that a sensor is working properly. Thus, in other words,

type I errors occur when a correct sensor is reported as faulty while type II errors occur when faulty sensors are not detected.

The criteria for deciding if a reading is faulty or not can result in a trade off between these two types of errors. The criteria considered in this project are the following:

1. Calculate the distance of the real value from the expected value, and map it to faulty if it is beyond a specified threshold and to correct if it is less than a specified threshold. The threshold values considered were 2, 2.5 and 3 times the standard deviation $\sigma$.

2. Assume that the sensor is working properly and establish a confidence level at which this hypothesis can be rejected, in which case it can be considered faulty. This confidence level is known as the $p$ value. The $p$ values considered were 0.05 and 0.01.

The accuracy of the model, i.e., the proportion of type I and II errors, is evaluated by varying the possible thresholds for each of these criteria.

Two different faults were simulated:

**Severe.** The sensor value modified is the most distant extreme value, i.e., the real value is substituted by one which differs by minimum 50 %.

**Mild.** The real value is replaced by one which differs by 25 %.

A test procedure was used to evaluate the accuracy of the whole validation process. Table 2 presents the final evaluation of the prototype with the percentage of type I and II errors for severe and mild faults.

Table 2: Final evaluation: number of errors and their percentage for severe and mild faults.

| Criteria | $2\sigma$ | $2.5\sigma$ | $3\sigma$ | $p = 0.05$ | $p = 0.01$ |
|---|---|---|---|---|---|
| Severe fault | | | | | |
| Type I | 17.3 % | 4.8 % | 2.9 % | 14.5 % | 2.9 % |
| Type II | 0.5 % | 0.1 % | 0.9 % | 0.7 % | 0.4 % |
| Mild fault | | | | | |
| Type I | 21.5 % | 10.3 % | 11.4 % | 17.7 % | 6.5 % |
| Type II | 8.0 % | 11.7 % | 14.9 % | 4.7 % | 5.8 % |

Type I errors imply that most of the sensors in a EMB present apparent type I errors. This is more common as it can be seen in Table 2. That is, there are cases where the existence of an invalid apparent fault, together with the valid ones, completes the EMB of a misdiagnosed sensor. Hence, a type I error is produced. On the contrary, type II errors are detected at this stage when most of the sensors of a EMB present misdiagnosed apparent faults. This is very improbable as the results of Table 2 confirms. The percentages are obtained comparing the average number of errors, with the total number of experiments.

# 6 Discussion

Section 4.2 developed an any time sensor validation algorithm that utilizes an entropy function as a criterion for selecting the next sensor to validate. This entropy function calculates the amount of information that any single validation provides for diagnosing all the sensors. Hence, to evaluate this criterion, this section compares the performance profile of the any time sensor validation algorithm as a function of time when the entropy based measure is used, and when a random selection scheme is used.

Figure 11 shows the resultant performance profile of the any time sensor validation algorithm. That is, the quality of the response as a function of time. An experiment consisted in the simulation of a single fault. Thus, 21 independent experiments were necessary to simulate a fault in all the sensors. In total, 260 experiments were carried out, so every one of the 21 sensors was simulated faulty at least 12 times (12.6 times). The *entropy* graph represents the average of the resultant quality with the entropy based scheme for the 21 sensors of Fig. 10. The *random* graph represents the average of the same experiment with a random selection scheme. The time axis is a qualitative comparison rather than quantitative.

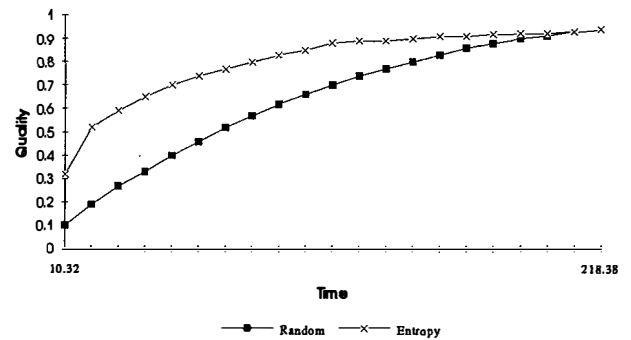

Figure 11: Performance profile of the any time sensor validation algorithm ($time \times 10^{-2}sec.$).

Alternatively, the results can also be evaluated by comparing the time required to reach different levels of quality. For example in Fig. 11, when the random criterion reaches 60 % of quality, the entropy criterion has already reached more than 80 %.

The approach has been implemented and is being tested on the validation of temperature sensors in a gas turbine of a combined cycle power plant. The results for the accuracy of the model were reported in terms of the type I and type II errors and with respect to detecting severe and mild faults. The results showed, that for this particular test application, more stringent criteria for detecting failures reduced type I errors but did not significantly increase type II errors.

The results of the evaluation of the validation and isolation phases together are shown in Table 2. Again, with a *p value* of 0.01, there are 2.9 % of type I errors, and 0.4 % type II errors. Notice that, in general, the sensor validation algorithm performs almost perfectly with respect to undetected faulty sensors, i.e., all the faults are detected. At the same time, the rate of incorrect detection faults is satisfactory for most of the criteria analyzed.

Two complexity aspects need to be discussed. The first one is the size of the pre compiled decision tree presented in section 4.2. A binary tree for $n$ sensors contains $n$ levels and up to $2^{n-1}$ nodes (1,048,575 nodes for 21 sensors). However, if a single fault is assumed, then the decision tree results in a pruned tree with $n$ levels and at most $n \times (n+1)$ nodes. The second one is the complexity for probability propagation in the fault isolation network as in Fig. 8. The propagation complexity (using the clostering algorithm [Lauritzen & Spiegelhalter 1988]), depends on the the size of the largest entry of the $EMB$ table, i.e., the largest clique. However, if a tree is assumed for the detection Bayesian network, the number of nodes in the $EMB$ table remains small, i.e., just one parent and the children of a node.

## 7   Conclusions

This paper has presented an any time, probabilistic algorithm for sensor validation. A layered approach is considered where the lowest layer performs the validation. A Bayesian network is used to define the relationships between variables and to estimate the expected value of a sensor. The expected value is then compared with the actual reading obtained. If these measures differ then a faulty sensor is suspected. A faulty sensor is then distinguished from apparently faulty sensors by the use of a property based on the Markov blanket.

An any time version of the validation algorithm, that improves the quality of its answer incrementally, has also been presented. This any time algorithm uses a causal network to distinguish the real fault from the apparent ones. The any time behaviour is obtained with the selection of the sensor that provides more information when validated. The selection is made with the entropy function.

The evaluation of the any time behaviour of the algorithm presented in this paper was done by carrying out experiments to obtain the performance profile of the entropy based selection scheme and comparing it with a random selection scheme.

Future research will attempt to use the probabilities obtained in the fault detection Bayesian network, as the input to the fault isolation Bayesian network. At this stage, the output of the detection network consists in a binary value ($\{correct, faulty\}$).

## Acknowledgments

Thanks to the anonymous referees for their comments which improved this article. This research is supported by a grant from CONACYT and IIE under the IIE/SALFORD/CONACYT doctoral programme.